\pdfoutput=1
\documentclass[11pt]{article}

\usepackage{EACL2023}

\usepackage{times}
\usepackage{latexsym}

\usepackage[T1]{fontenc}
\usepackage[utf8]{inputenc}

\usepackage{microtype}

\usepackage{inconsolata}

\usepackage{multirow}
\usepackage{adjustbox}
\usepackage{color}
\usepackage{caption}
\usepackage{subcaption}
\usepackage{bbm}
\usepackage{booktabs}
\usepackage{amsmath}
\usepackage{amssymb}
\DeclareMathOperator*{\argmax}{argmax}

\newcommand{\scriptD}{\mathcal{D}}
\newcommand{\scriptX}{\mathcal{X}}
\newcommand{\scriptY}{\mathcal{Y}}

\newcommand{\scriptL}{\mathcal{L}}
\usepackage{amsmath}
\usepackage{nccmath}

\usepackage{enumitem}
\setlist[itemize]{align=parleft,left=0pt..1em}
\usepackage{xcolor,pifont}
\newcommand*\colourcheck[1]{%
  \expandafter\newcommand\csname #1check\endcsname{\textcolor{#1}{\ding{52}}}%
}
\colourcheck{blue}
\colourcheck{green}
\colourcheck{red}
\definecolor{darkgreen}{rgb}{0.3, 0.7, 0.0}

\title{Bag of Tricks for In-Distribution Calibration of Pretrained Transformers}

\author{Jaeyoung Kim \\
  VUNO, Inc. \\
  \texttt{jaeyoung.kim@vuno.co} \\\And
  Dongbin Na \\
  VUNO, Inc. \\
  \texttt{dongbin.na@vuno.co} \\ \And
  Sungchul Choi \\
  Pukyong National University \\
  \texttt{sc82.choi@pknu.ac.kr} \\
  \AND
  Sungbin Lim\thanks{Corresponding author. This work is partially done at UNIST.} \\
  Korea University \\
  \texttt{sungbin@korea.ac.kr} \\}

\begin{document}
\maketitle
\begin{abstract}
While pre-trained language models (PLMs) 
have become a de-facto standard promoting the accuracy of text classification tasks, recent studies~\cite{MISCALIBRATE,dan2021effects} find that PLMs often predict over-confidently.
Although various calibration methods have been proposed, such as ensemble learning and data augmentation, most of the methods have been verified in computer vision benchmarks rather than in PLM-based text classification tasks.
In this paper, we present an empirical study on confidence calibration for PLMs, addressing three categories, including confidence penalty losses,
data augmentations, and ensemble methods.
We find that the ensemble model overfitted to the training set shows sub-par calibration performance and also observe that PLMs trained with confidence penalty loss have a trade-off between calibration and accuracy.
Building on these observations, we propose the \textbf{Cal}ibrated P\textbf{L}M (CALL), a combination of calibration techniques.
The CALL complements the drawbacks that may occur when utilizing a calibration method individually and boosts both classification and calibration accuracy.
Design choices in CALL's training procedures are extensively studied, and we provide a detailed analysis of how calibration techniques affect the calibration performance of PLMs.
\end{abstract}

\section{Introduction}

Trustworthy deployment of machine learning applications requires accurate and calibrated predictions to instill their reliability and help users be less confused about models' decisions~\cite{NLPUQ1,SNGP}.

\begin{figure}[th]
  \centering
    \begin{subfigure}[b]{0.23\textwidth}    
         \centering
         \includegraphics[width=\textwidth]{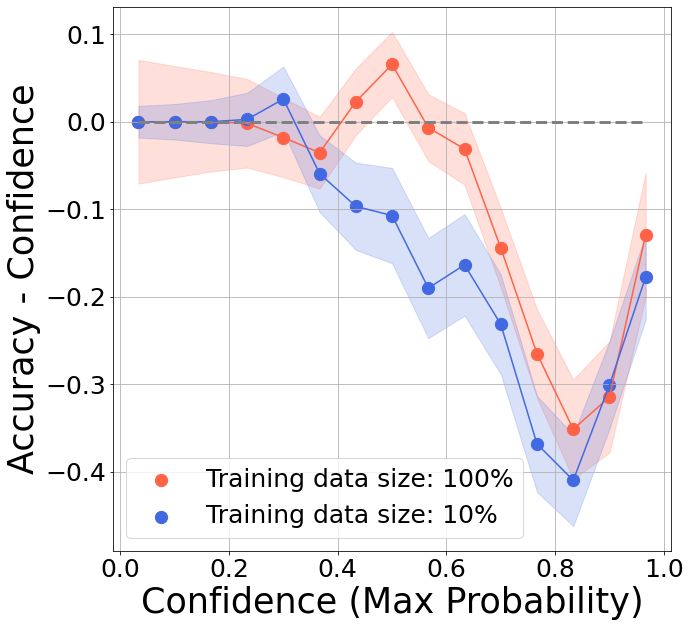}
         \subcaption{BERT}
     \end{subfigure}
    %  \hfill
     \begin{subfigure}[b]{0.23\textwidth}
         \centering
         \includegraphics[width=\textwidth]{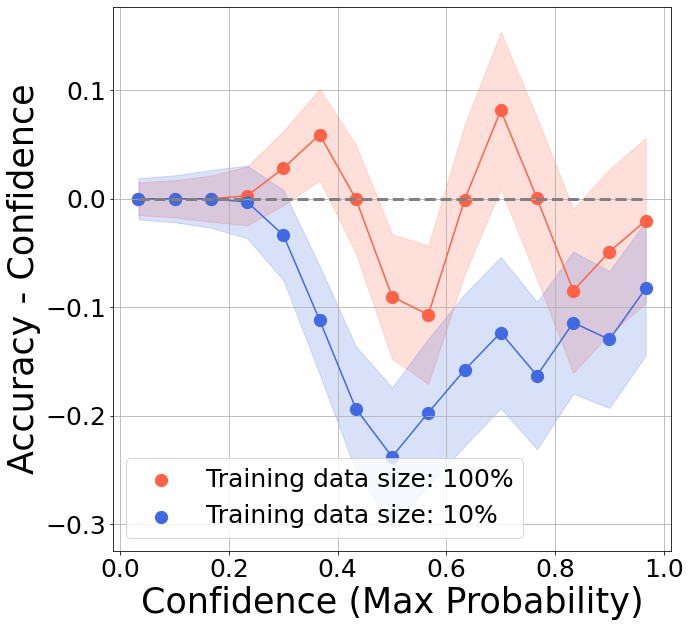}
         \subcaption{RoBERTa}
     \end{subfigure}
  \caption{Reliability diagrams \cite{diagram} on TREC \cite{TREC} with PLMs. A dashed line implies a perfect calibration while 
  PLMs generally show over-confident predictions.
  }
  \label{fig:intro_diagram}
\end{figure}

However, modern deep neural networks (DNNs) produce miscalibrated predictions, i.e., a mismatch between a model's confidence and its correctness.
One of the reasons is that an over-parameterized classifier typically produces over-confident predictions \cite{ce_overconfident}.
Moreover, the miscalibration can be exacerbated when DNNs make predictions on test data different from the training distribution, i.e., distribution shift \cite{distribution_shift_1}.

To obtain the well-calibrated predictions, many pioneering studies have shown the calibration effect of ensemble and regularization techniques focused on computer vision benchmarks.
Ensemble learning has become one of the standard approaches to reduce calibration errors \cite{DE_origin,bonab2019less}.
\citet{ERL} propose the entropy regularized loss which penalizes confident output distributions in order to reduce overfitting.
\citet{MIXUP,AUGMIX} demonstrate that DNNs trained on diverse augmented data are less prone to produce over-confident predictions, leading to the calibration benefit under the distribution shift.

Intense research effort has focused on improving the calibration performance of vision models on image datasets.
However, exploration of existing calibration methods with pre-trained Transformers (PLMs) has received less attention.
Moreover, recent studies show that PLMs such as BERT~\cite{BERT} and RoBERTa~\cite{ROBERTA} produce miscalibrated predictions introduced by over-parameterization~\cite{MISCALIBRATE}.
Therefore, it is necessary to investigate how modern calibration techniques affect PLMs' calibration.

In this paper, focused on PLMs in multi-class classification tasks, we explore widely used calibration families, including (1) confidence penalty loss functions that can be used instead of cross-entropy loss, (2) data augmentations, and (3) ensemble methods.
We consider a low-resource regime since the small size of the training dataset amplifies the miscalibration of models~\cite{data_scarcity_1}.
We also observe PLMs especially produce unreliable predictions in the data scarcity setting (see Figure \ref{fig:intro_diagram}).

\noindent \textbf{Contributions.}
We conduct a comprehensive empirical study for the effectiveness of the above calibration methods. In this study, our findings are as follows:
\begin{itemize}
    \item A PLM trained with imposing a strong penalty on the over-confident output shows significant improved calibration performance, but its accuracy can slightly deteriorate.
    \item For ensemble methods, Deep Ensemble~\cite{DE_origin} and MIMO~\cite{MIMO} increase the diversity of predictions, resulting in the well-calibrated predictions in the data scarcity setting.
    However, the ensemble methods show insufficient calibration when each ensemble member is overfitted to negative log-likelihood for the training dataset.
    \item Data augmentation methods that can expose diverse patterns such as MixUp~\cite{MIXUP} and EDA~\cite{EDA} are more effective for calibration in PLMs compared to weak text-augmentation methods~\cite{SR,AEDA}.
\end{itemize}

Building on our findings, we present Calibrated PLM (CALL), a blend of the discussed calibration methods. Numerical experiments demonstrate that the components of CALL complement each other's weaknesses. For instance, data augmentation and ensemble methods offset the accuracy decline caused by the confidence penalty loss, while data augmentation and the confidence penalty loss counteract overfitting in the ensemble model.
Through our extensive experiments, we show the CALL’s competitiveness on several text classification benchmarks.

\section{Related Work}

The calibration of machine learning models has been mainly studied for the trustworthy deployment of image recognition applications \cite{DE_origin,MIXUP,ce_overconfident}.
Beyond the computer vision fields, research on the calibration ability of language models in the NLP domain has also recently been attracting attention \cite{PLM_CAL_01,dan2021effects}.

\citet{PLM_CAL_01} investigate the calibration ability of PLMs, and they demonstrate that RoBERTa produces more calibrated predictions than BERT. They also show that temperature scaling \cite{TEMPERATURE} and label smoothing \cite{LS} improve the calibration performance of PLMs for language understanding tasks.
\citet{dan2021effects} conduct an empirical study of the effects of model capacity on PLMs and show that smaller pre-trained transformers provide more reliable predictions.
\citet{MASKER} find that PLMs tend to produce over-confident outputs based on in-distribution (ID) keywords rather than contextual relations between words. 
They demonstrate that keyword-biased predictions can be over-confident even in out-of-distribution samples with ID keywords.

\citet{MISCALIBRATE} suggest two regularizers using generated pseudo-manifold samples to improve both ID and out-of-distribution calibration for PLMs. They use MixUp \cite{MIXUP} as a regularization technique for BERT calibration and show that mixed training samples on the data manifold improve the calibration performance.
Similarly, \citet{park-caragea-2022-calibration} propose a variant of MixUp utilizing saliency signals and also analyze the impact of combining additional calibration methods with MixUp. However, they only consider temperature scaling and label smoothing as additional calibration methods.

\section{Why Re-assess Calibration Methods?}
\citet{ce_overconfident} observe that a larger DNN tends to be more poorly calibrated than a smaller one.
As the size of the parameters for modern DNNs continues to increase, the miscalibration issues need to be addressed more than ever.

At the same time, the unique character of PLMs raises concerns about whether previous findings on calibration obtained from standard convolutional neural networks (CNNs) can be successfully extended to PLM.
For example, PLMs with ensemble learning may have different behavior compared to randomly initialized CNNs because naive PLMs have a massive amount of parameters and are initialized with pre-trained weights in the fine-tuning stage.

On the other hand, for the data augmentation, because image transformations (e.g., flipping, translation, and rotating) can not be directly applied to text-based samples, thus, it is also necessary to investigate the effect of text-specific augmentations on the calibration of PLMs.

\section{Calibration Strategies}
\label{sec:method}

In this section, we review the existing literature used in our experiments and how we applied each method to PLMs. Calibration methods we explore are denoted by \textbf{bold}.

\subsection{Preliminaries}

\noindent \textbf{Notation.}
Let $\scriptD=\{x_i,y_i\}_{i=1}^N$ be a dataset consisting of $N$ samples, where $x_i \in \scriptX$ is a input and $y_i \in \scriptY = \{1,...,K\}$ is a ground truth label.
We denote by $\bar{p}_{i}=f(y|x_i)$ the predicted distribution of a classifier $f$.
Class prediction and associated confidence (maximum probability) of $f$ are computed as $\hat{y}_i = \argmax_{k \in \scriptY} \bar{p}_{i}$ and $\hat{p}_i = \max_{k \in \scriptY} \bar{p}_{i}$, respectively.

In the BERT-style architecture, output of embedding layer, $L$ attention blocks, and the output dense layer (with softmax function) are denoted by $z_{\text{embed}}$, $g = \{g_1,...,g_L\}$, and $h$, respectively.

\noindent \textbf{Calibration Metrics.} A calibrated model provides reliable predictive probability whose confidence aligns with its expected accuracy, i.e. $\mathbb{E}_{\hat{p}}[|\mathbb{P}(\hat{y}=y|\hat{p})-\hat{p}|]$.
Given a finite dataset, \textit{Expected Calibration Error} (ECE; \citealp{ECE}) is widely used as a calibration performance measure. ECE can be computed by binning predictions into $T$ groups based on predictions of $f$ and then taking a weighted average of each group's accuracy/confidence difference:

\begin{equation}\label{eq:ece}
    \sum_{t=1}^T \frac{|B_t|}{N} |\text{acc}(B_t) - \text{conf}(B_t)|,
\end{equation}

where $B_t$ is the group of samples and their corresponding confidences belonging to the $(\frac{t-1}{T}, \frac{t}{T}]$.
The acc($B_t$) and conf($B_t$) denote average accuracy and confidence of predictions for $B_t$, respectively.

Model calibration also can be measured using proper scoring rules \cite{proper_scoring} such as Brier score \cite{brier} and negative log likelihood (NLL).

\subsection{Confidence Penalty Losses}

We explore an alternative loss functions  that can be used instead of cross-entropy (CE) loss.

\noindent \textbf{Brier Loss} (BL; \citealp{brier}) is one of the proper scoring rules, defined as the squared error between the softmax output and the one-hot ground truth encoding. BL is related to ECE in that it is an upper bound of the calibration error by the calibration-refinement decomposition \cite{BL_reference,SNGP}.

\noindent \textbf{Entropy Regularized Loss} (ERL; \citealp{ERL}) penalizes confident output distributions by adding the negative entropy:
\begin{equation}
    \scriptL_{\text{ERL}} = \scriptL' + \beta \sum_{k=1}^K \bar{p}_{k} \log \bar{p}_{k},
\end{equation}
where $\scriptL'$ can be an arbitrary classification-based objective function (e.g., CE and BL), and $\beta$ is the hyperparameter that controls the strength of the confidence penalty.

\noindent \textbf{Label Smoothing} (LS; \citealp{LS}) is a commonly used \textit{trick} for improving calibration that generates a soft label by weighted averaging the uniform distribution and the hard label.

\subsection{Data Augmentations}

Data augmentations have been widely used to improve the model's calibration performance in computer vision fields \cite{MIXUP,AUGMIX,AUGMAX}.
However, text augmentations are often overlooked in the literature on the calibration in NLP tasks.
To the best of our knowledge, we are the first to extensively study how text augmentation techniques such as Synonym Replacement (SR; \citealp{SR}), Easy Data Augmentation (EDA; \citealp{EDA}), and An Easier Data Augmentation (AEDA; \citealp{AEDA}) affect calibration performance.
We also investigate the recent variant of MixUp \cite{MIXUP_CLS}.

\noindent \textbf{SR} randomly choose $n$ words from the input sentence except for stop words and then replace each of these words with one of its synonyms chosen using WordNet \cite{WORDNET}.

\noindent \textbf{EDA} is a token-level augmentation method that consists of four random transformations: SR, Random Deletion, Random Swap, and Random Insertion. 

\noindent \textbf{AEDA} only use Random Insertion operator that insert punctuation marks (i.e., ``.'', ``,'', ``!'', ``?'', ``;'', ``:'') into a input sentence.

\noindent \textbf{MixUp} \cite{MIXUP} is a data augmentation strategy using convex interpolations of inputs and accompanying labels.
\citet{MIXUP_TEXT} investigate word- and sentence-level MixUp strategies to apply MixUp to recurrent neural networks.
\citet{MIXUP_CLS} propose MixUp-CLS, that performs MixUp on the pooled \verb|[CLS]| token embedding vector for a last attention layer of PLM. MixUp-CLS shows improved accuracy for natural language understanding (NLU) tasks compared to word-level MixUp. Unless otherwise specified, we use MixUp-CLS in our experiment.

\subsection{Ensembles}

Ensemble techniques utilize $M$ models by combining them into an aggregate model and then average the predictions to produce calibrated outputs: $ \frac{1}{M} \sum_{m=1}^M f_m(y|x)$.
We compare the deterministic model with three ensemble approaches, and the computational cost of the ensemble methods used in the experiment is reported in Appendix \ref{sec:computational_cost}.

\noindent \textbf{Deep-Ensemble} (DE; \citealp{DE_origin}) consists of $M$ randomly initialized models and provides a calibration effect leveraging the predictive diversity of ensemble members.
When applying DE to PLMs, $M$ independent models have different initialization weights only in a penultimate layer since PLMs are initialized with pre-trained weights.

\noindent \textbf{Monte Carlo Dropout} (MCDrop; \citealp{MCDrop}) interprets Dropout as an ensemble model, leading to its application for uncertainty estimates by sampling $M$ times dropout masks at test time.

\noindent \textbf{Multi-Input and Multi-Output (MIMO).}
To alleviate the high computational cost and memory inefficiency of DE, \citet{MIMO} propose the multi-input and multi-output architecture by training $M$ sub-networks inside a CNN.

In original MIMO, the $M$ inputs (images) $\{x^m\}_{m=1}^{M}$ are sampled from $\scriptD_{\text{train}}$. 
MIMO concatenates multiple inputs per channel before the first convolution layer and produces multiple outputs using $M$ independent output dense layers. The feature extractor of CNN remains unchanged.
For the training procedure, all ensemble members have the same mini-batch inputs with probability $p$, and the inputs are randomly sampled from the training dataset with probability $1-p$.

For applying MIMO to the PLMs, the following consideration arise; 
When multiple inputs are connected before the embedding layer, the length of tokens is $M$ times longer.
Thus, applying MIMO to PLMs in this manner is inefficient for a dataset that consists of long sentences.

Instead, we modify the original configuration of MIMO so that it can be applied to various NLP tasks.
For PLM, the output of the first attention layer $\bar{z}$ is calculated by averaging multiple outputs of $M$ independent first attention blocks $\{g_{\text{1}}^m\}_{m=1}^M$:
\begin{equation}
    \bar{z} = \frac{1}{M} \sum_{m=1}^M \, g^m_{\text{1}}(z_{\text{embed}}).
\end{equation}

To produce multiple predictions, we use $M$ modules that consist of the last attention blocks $\{g_\text{L}^m\}_{m=1}^M$ and dense layer $h$. 
The ensemble prediction is calculated by:
\begin{equation}
    \bar{p} = \frac{1}{M} \sum_{m=1}^M h(g^m_{\text{L}}(g'(\bar{z})),
\end{equation}
where $g' = \{g_2,...,g_{L-1}\}$ is the shared attention blocks.

\begin{table}[h]
\centering
\begin{adjustbox}{width=7.0cm,center}
\begin{tabular}{c|ccccc} \hline
 & \# train & \# dev & \# test & $l_{avg}$ & \# classes \\ \hline
SST2 & 7.0k & 0.7k & 1.8k & 19 & 2\\
20NG & 9.1k & 2.2k & 7.5k & 320 & 20\\
TREC & 4.9k & 0.5k & 0.5k & 10 & 6 \\
\hline
\end{tabular}
\end{adjustbox}
\caption{Summary of data statistics. $l_{avg}$: Sentence average length.}
\label{tab:data_statistics}
\end{table}

\begin{table*}[h]
\begin{adjustbox}{width=12.5cm,center}
\begin{tabular}{l|ccc} \toprule
Acc$\uparrow$ / ECE$\downarrow$ / NLL$\downarrow$ & \textbf{TREC} & \textbf{SST2} & \textbf{20NG} \\ \hline
RoBERTa (baseline) & \underline{94.04} / 4.08 / 24.86 & \underline{91.23} / 7.42 / 43.08  & 76.58 / 11.37 / 90.40 \\ 
CE+ERL & 93.72 / 4.05 / 24.20 & 91.04 / 6.62 / 38.77 & \underline{76.79} / 11.21 / 90.32 \\
CE+LS & 93.84 / 3.37 / \underline{23.71} & 91.16 / 6.03 / 30.26 & 76.39 / 11.36 / 90.90 \\
BL & 93.24 / 2.69 / 26.55 & 89.48 / 7.15 / 36.02 & 75.74 / 7.21 / \underline{86.02} \\
BL+ERL & 93.84 / 2.48 / 24.78 & 90.32 / 5.68 / 29.61 & 76.13 / 6.62 / 86.11 \\
BL+LS & 93.52 / \underline{2.32} / 25.16 & 91.15 / \underline{5.56} / \underline{29.37} & 75.83 / \underline{6.57} / 86.31 \\
\hline
SR & 94.24 / 3.37 / 22.24 & 90.54 / 7.22 / 38.03 & 76.45 / 10.54 / \underline{87.64} \\
AEDA & 93.76 / 4.68 / 28.36 & \textcolor{black}{\textbf{91.45}} / 6.69 / 37.67 & 76.41 / 11.49 / 91.21 \\
EDA & 93.40 / 2.83 / 23.46 & 91.56 / \underline{5.01} / \underline{29.86} & 76.01 / \underline{10.52} / 88.89 \\
MixUp & \underline{94.76} / \textcolor{black}{\textbf{2.23}} / \underline{22.02} & 90.86 / 6.46 / 31.89 & \underline{76.74} / 11.22 / 90.65 \\
\hline
MCDrop & 94.20 / 4.16 / 24.45 & 91.04 / 6.84 / 39.55 & 76.63 / 10.18 / 87.52 \\
MIMO & 94.88 / 3.13 / 20.38 & 91.26 / 6.21 / 32.78 & 76.25 / \textcolor{black}{\textbf{5.61}} / 81.43 \\
DE & \textcolor{black}{\textbf{95.03}} / \underline{2.89} / \textcolor{black}{\textbf{19.02}} & \underline{91.44} / \textcolor{black}{\textbf{4.88}} / \underline{29.51} & \textcolor{black}{\textbf{78.09}} / 7.51 / \textcolor{black}{\textbf{78.96}} \\

\bottomrule 
\end{tabular}
\end{adjustbox}
\caption{Results for the low-resource regime. For each dataset, all methods are trained with $10\%$ of  training samples. The best results in each category are indicated in \underline{underline} and the best results among all methods are indicated in \textbf{bold}.
Accuracy is a percentile. We report ECE and NLL multiplied by $10^2$.
}
\label{tab:low_resource_result}
\end{table*}

\section{Experiments}
\label{sec:results}
This section presents the experimental results of the calibration methods. We describe experimental datasets and settings (Section \ref{sec:dataset} and \ref{sec:setting}), followed by empirical results for the low-resource regime (Section \ref{sec:low_resource_result}), overall calibration result (Section \ref{sec:overall_result}), and detailed analysis (Section \ref{sec:analysis}).
We then introduce the training procedure of CALL in Section \ref{sec:call}.
In our experiments, we set RoBERTa trained with CE as a baseline.
Unless otherwise specified, ensemble and augmentation methods are applied to the baseline.

\subsection{Datasets and Metrics}
\label{sec:dataset}
\noindent \textbf{Dataset}. Following \citet{CONTRA_NLP}, we use the following three text classification datasets. Data statistics are described in Table \ref{tab:data_statistics}.

\begin{itemize}
    \item Stanford Sentiment Treebank (SST2; \citealp{SST2}) is a sentiment analysis dataset that consists of sentences from movie reviews.
    \item 20 Newsgroups (20NG; \citealp{20news}) is a topic categorization dataset which contains news articles with 20 categories.
    \item TREC \cite{voorhees-tice-2000-trec} is a dataset for question classification, and we use its coarse version with six classes.
\end{itemize}
To evaluate the effectiveness for calibration methods in the data scarcity setting, we use 10\% of the training set.

\noindent \textbf{Metrics}. We measure ECE and NLL for each calibration method. For ECE, we bin the predictions into $T=15$ equidistant intervals.
We report ECE and NLL multiplied by $10^2$ in all experimental results for the convenience.

\subsection{Training Configurations}
\label{sec:setting}

We implement our framework upon Huggingface’s Transformers \cite{HUGGINGFACE} and build the text classifiers based on RoBERTa (\verb|roberta-base|) in the main experiment.
All models are optimized with Adam optimizer~\cite{ADAM} with a weight decay rate of $0.01$, warmup proportion of $0.1$, batch size of 16, a dropout rate of $0.1$, and an initial learning rate of $\text{1e-5}$.
We fine-tune the RoBERTa for $10$ epochs.
For each calibration method, hyper-parameters are tuned according to the classification performance, and the detailed hyper-parameter setting is described in Appendix \ref{sec:hyperparameter_setting}.
We also provide empirical results for BERT (\verb|bert-base-cased|) in Appendix \ref{sec:bert_result}.
We report the averaged performance over 5 runs using different random seeds and implementation results are available at \url{https://github.com/kimjeyoung/PLM_CALL}.

\begin{table*}[h]
\begin{adjustbox}{width=12.5cm,center}
\begin{tabular}{l|ccc} \toprule
Acc$\uparrow$ / ECE$\downarrow$ / NLL$\downarrow$ & \textbf{TREC} & \textbf{SST2} & \textbf{20NG} \\ \hline
RoBERTa (baseline) & \underline{97.40} / 2.41 / 15.24 & 94.35 / 4.13 / 26.36  & 86.00 / 9.51 / 68.26 \\ 
CE+ERL & 97.24 / 2.44 / 14.64 & 94.05 / 4.05 / 26.94 & 86.13 / 9.41 / 70.18 \\
CE+LS & 97.28 / 2.06 / 13.11 & 94.21 / 3.75 / 20.17 & 86.14 / 9.81 / 70.16 \\
BL & 97.04 / 1.80 / 12.23 & 94.48 / 2.95 / \textcolor{black}{\textbf{17.25}} & 86.06 / 7.06 / 58.37 \\
BL+ERL & 97.28 / \textcolor{black}{\textbf{1.35}} / \underline{12.09} & \textcolor{black}{\textbf{94.97}} / 3.21 / 17.31 & 85.77 / \textcolor{black}{\textbf{6.75}} / \textcolor{black}{\textbf{58.02}} \\
BL+LS & 96.92 / 1.41 / 12.54 & 94.34 / \underline{2.78} / 17.74 & \underline{86.15} / 6.76 / 58.17 \\
\hline
SR & 97.04 / 2.19 / 12.18 & 94.31 / 3.48 / 20.81 & 85.97 / 9.31 / 64.84 \\
AEDA & \underline{97.24} / 2.35 / 12.99 & 94.45 / 3.70 / 23.27 & 85.89 / 9.85 / 69.41 \\
EDA & 97.16 / 1.87 / \textcolor{black}{\textbf{11.54}} & 94.21 / \underline{2.95} / 19.27 & 85.74 / \underline{8.69} / \underline{60.90} \\
MixUp & 97.20 / \underline{1.55} / 11.58 & \underline{94.57} / 3.61 / \underline{19.04} & \underline{86.21} / 8.72 / 64.48 \\
\hline
MCDrop & \textcolor{black}{\textbf{97.56}} / 2.37 / 13.84 & 94.01 / 3.64 / 24.02 & 85.97 / 8.61 / 64.49\\
MIMO & 97.32 / 2.30 / 12.86 & 94.32 / \textcolor{black}{\textbf{2.68}} / \underline{17.51} & 85.80 / 8.68 / \underline{60.92} \\
DE & 97.32 / \underline{2.09} / \underline{12.83} & \underline{94.64} / 3.10 / 19.15 & \textcolor{black}{\textbf{86.81}} / \underline{7.90} / 62.31 \\

\bottomrule 
\end{tabular}
\end{adjustbox}
\caption{Overall calibration results for calibration techniques. For each dataset, all methods are trained with $100\%$ of training samples.}
\label{tab:overall_result}
\end{table*}

\subsection{Result for Low-resource Regime}
\label{sec:low_resource_result}

Table \ref{tab:low_resource_result} represents the classification accuracy and calibration performances for each dataset in the low-resource regimes.
Most calibration strategies perform better than the baseline, even in cases where the baseline calibration results were already good, e.g., TREC.
These results demonstrate that the existing methods can enhance PLM's calibration ability when the annotation budget is small, as in many real-world settings.

Interestingly, augmentation methods except for AEDA also result in the calibration benefit. For example, MixUp and EDA show improved calibration performances for all datasets compared to the baseline.

Among confidence penalty losses, BL significantly reduces ECE for the three datasets. Moreover, the calibration performance is further improved when BL is combined with an additional regularization method (i.e., BL+ERL and BL+LS).
However, BL+LS and BL+ERL underperform the baseline with respect to accuracy, and this performance drop is also observed when applied to BERT (Appendix \ref{sec:bert_result}).

DE not only shows the most remarkable improvement of NLL but also improves accuracy for all datasets.
MIMO also consistently outperforms the baseline for ECE.
In summary, DE and MIMO are more effective than the other calibration methods when considering both accuracy and calibration in the low-resource regime.

\begin{figure*}[th]
  \centering
  \includegraphics[width=13.0cm]{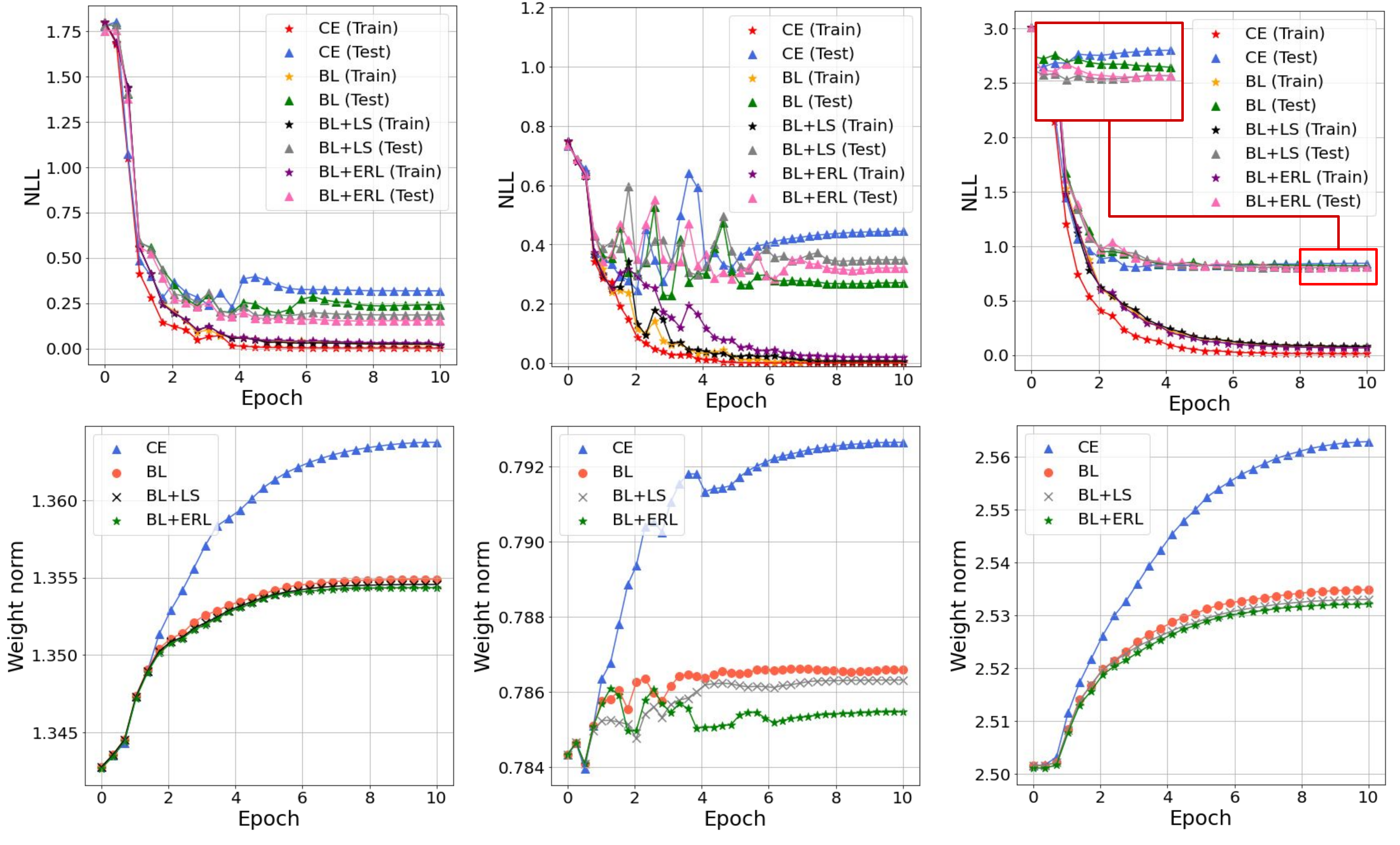}
  \caption{The plot of the NLL (Top) and the norm of weights (Bottom) while training RoBERTa on TREC (Left), SST2 (Middle), and 20NG (Right), respectively. The weights are extracted from the penultimate layer of RoBERTa and we use 10\% of samples for training. }
  \label{fig:nll_plot}
\end{figure*}

\subsection{Overall Result}
\label{sec:overall_result}

Overall performance result is reported in Table \ref{tab:overall_result}.
Similar to the results in Table \ref{tab:low_resource_result}, most of calibration methods show better calibration performance compared to the baseline.
In this setting, RoBERTa trained with BL+ERL works best.
For example, BL+ERL shows NLL results of 17.31 and 58.02 in SST2 and 20NG, respectively, but DE obtain 19.15 and 62.31.
In the data augmentation category, EDA and MixUp improve ECE and NLL compared to SR. AEDA underperforms the baseline for 20NG.

\subsection{Analysis}
\label{sec:analysis}

Our empirical results raise the following questions:
(1) Why do EDA and MixUp show better calibration performance than SR or AEDA?
(2) How can we improve the accuracy of BL+ERL?
(3) Why are ensemble methods more efficient than regularization methods in the low-resource setting, whereas BL+ERL is most effective for the full-data available setting?
We further conduct a detailed analysis focusing on the above questions.

\noindent \textbf{Role of Data Augmentation}. 
Although the PLM trained on the proper scoring rule reduce calibration error for the training dataset, minimizing calibration errors for all unseen ID samples is challenging because we use finite training data \cite{SNGP}.
As an alternative, if models trained with augmented samples learn diverse representations, we expect to match the distribution of training data with the distribution of unseen ID data.

\begin{table}[h]
\begin{adjustbox}{width=7.5cm,center}
\begin{tabular}{c|ccc} \toprule
Distance & \textbf{TREC} & \textbf{SST2} & \textbf{20NG} \\ \hline
SR & 11.56 / 17.44 & 7.12 / 12.01 & 15.54 / 23.02 \\
AEDA & 11.57 / 16.95 & 6.87 / 12.09 & 16.37 / 22.36 \\
EDA & 14.16 / 17.08 & \textbf{8.09} / \textbf{10.99} & \textbf{17.27} / 22.24 \\
MixUp & \textbf{14.52 / 15.44} & 7.69 / 11.18 & 16.65 / \textbf{21.62} \\

\bottomrule 
\end{tabular}
\end{adjustbox}
\caption{(Left) Distance between original and augmented sentences for the training samples. Higher is better. (Right) Distance between augmented training sentences and original test samples. Lower is better. The distance are computed at the last attention layer of RoBERTa.}
\label{tab:distance_exp}
\end{table}

\begin{table}[h]
\begin{adjustbox}{width=7.5cm,center}
\begin{tabular}{c|ccc} \toprule
Acc / ECE & \textbf{TREC} & \textbf{SST2} & \textbf{20NG} \\ \hline
BL+ERL & 93.84 / 2.48 & 90.32 / 5.68 & 76.13 / 6.62 \\
+SR & 92.60 / 2.93 & \textbf{91.75} / 4.57 & 76.40 / 5.49 \\
+AEDA & 93.84 / 2.84 & 91.32 / 5.21 & 75.83 / 6.14 \\
+EDA & 93.40 / 2.83 & 90.76 / 4.97 & \textbf{76.45} / \textbf{5.18} \\
+MixUp & \textbf{94.76} / \textbf{2.23} & 90.89 / \textbf{4.52} & 76.25 / 6.39 \\

\bottomrule 
\end{tabular}
\end{adjustbox}
\caption{Comparison result for augmentation methods. Each method is trained with 10\% of training data.}
\label{tab:BRERL_AUG}
\end{table}

We analyze the distance between unseen and training data distribution, assuming that the augmentation scheme that pulls the distribution of training data towards the unseen data distribution will be effective for calibration.

To measure the distance between the two distributions, we use Hausdorff-Euclidean distance.
In Table \ref{tab:distance_exp}, RoBERTa trained with MixUp shows the closest distance between training data and test data, followed by EDA.
In addition, the augmented data generated by MixUp and EDA are far away from the training data.
It can be interpreted that EDA and MixUp generate more diverse patterns of representations.
Hence, matching the distribution of observed data with the distribution of unseen data by adopting a proper augmentation method that generates diverse patterns may help the model produces calibrated predictions.

On the other hand, since data augmentation generally helps to improve accuracy, we investigate whether augmentation methods improve the accuracy of BL+ERL.
In Table \ref{tab:BRERL_AUG}, 
MixUp improves not only classification accuracy but also calibration performance on all datasets compared to the naive BL+ERL.

\noindent \textbf{Role of Regularization}. 
A crucial empirical observation by \citet{ce_overconfident} is that overfitting the NLL during training appears to be associated with the miscalibration of DNNs.

To better understand the role of strong regularization, we visualize the NLL during the training process of PLM.
In Figure \ref{fig:nll_plot}, training and test NLL are reduced at the beginning of training regardless of regularization methods.
However, as training progresses, the test NLL of RoBERTa trained with CE increases\footnote{Note that we use weight decay and dropout for training in order to alleviate overfitting.}.
On the other hand, other regularization methods show an inhibitive effect on overfitting compared to CE.

A DNN can produce over-confident predictions if the network increases the norm of its weights, which results in the high magnitudes of the logits \cite{FOCAL}. Figure \ref{fig:nll_plot} (Bottom) shows that the RoBERTa trained with CE also has a larger norm than the regularized models.

\begin{figure}[th]
  \centering
    \begin{subfigure}[b]{0.23\textwidth}
         \centering
         \includegraphics[width=\textwidth]{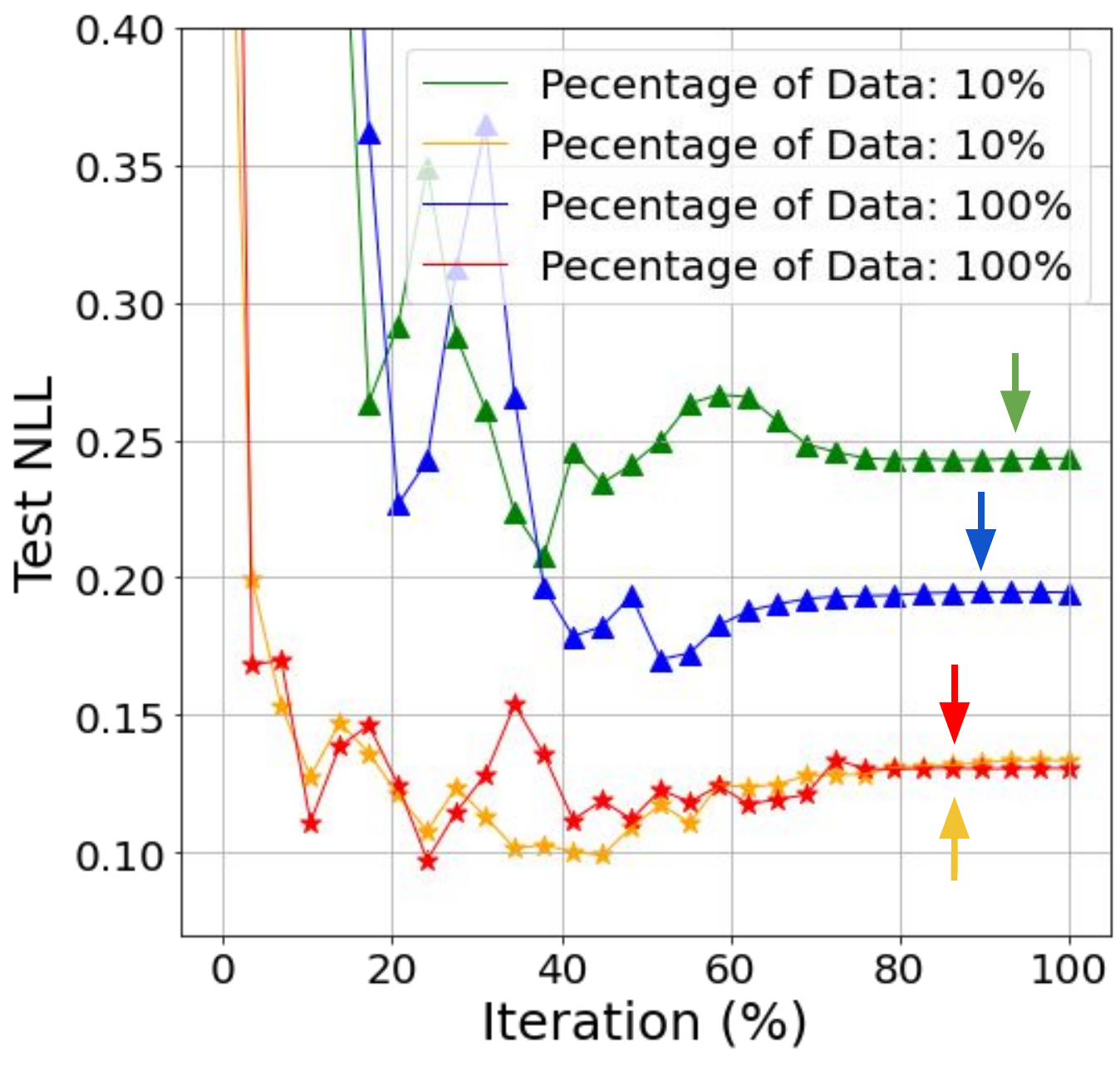}
         \subcaption{TREC}
     \end{subfigure}
    %  \hfill
     \begin{subfigure}[b]{0.23\textwidth}
         \centering
         \includegraphics[width=\textwidth]{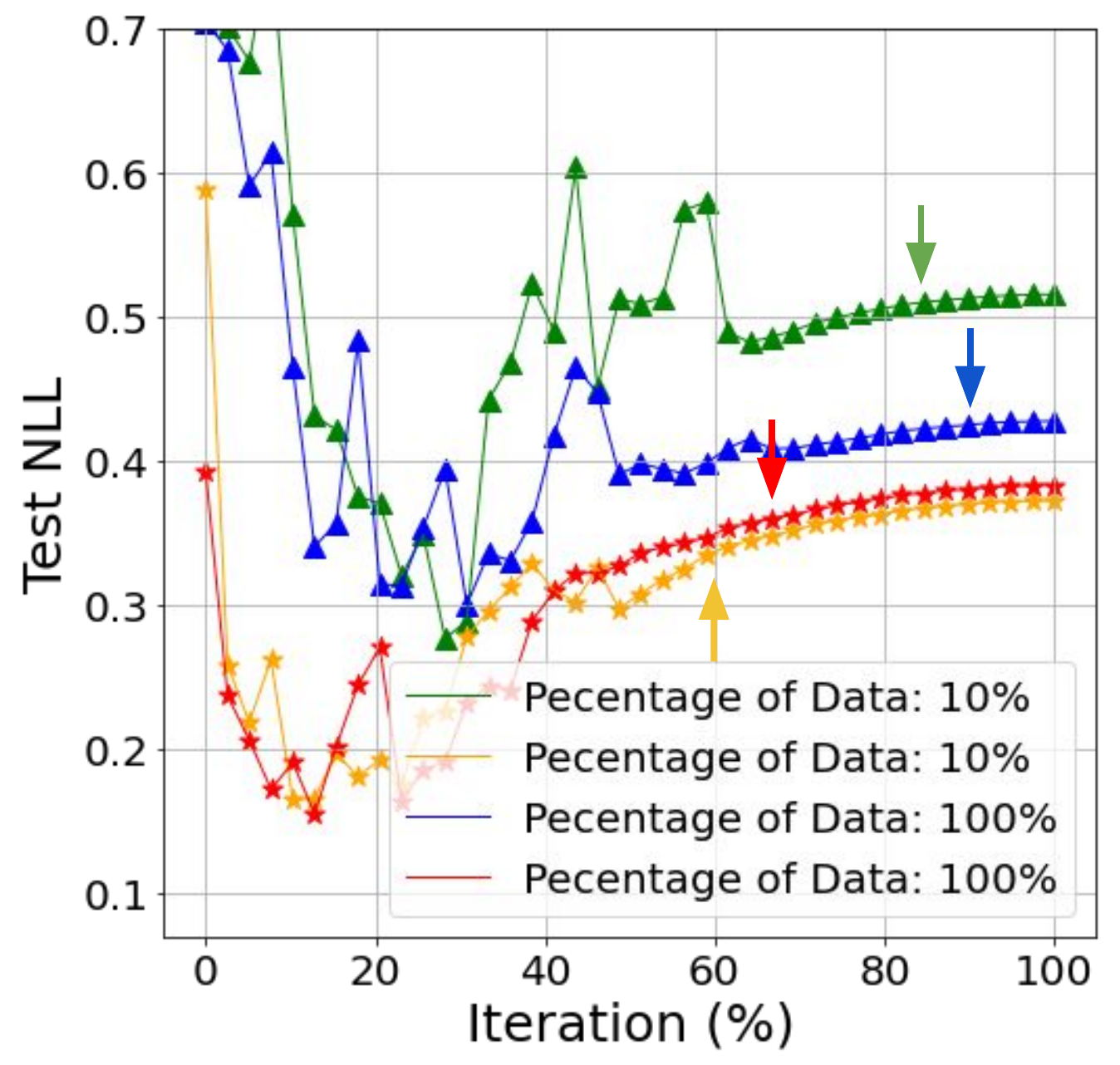}
         \subcaption{SST2}
     \end{subfigure}
  \caption{The test NLL for DE. Each arrow denotes the point at which the validation accuracy is the maximum.}
  \label{fig:de_nll}
\end{figure}

\begin{table*}[h]
\begin{adjustbox}{width=12.5cm,center}
\begin{tabular}{l|ccc} \toprule
Acc$\uparrow$ / ECE$\downarrow$ / NLL$\downarrow$  & \textbf{TREC} & \textbf{SST2} & \textbf{20NG} \\ \hline
Train samples &  & \textbf{100 \%} & \\ \hline
RoBERTa (baseline) & \underline{97.40} / 2.41 / 15.24 & 94.35 / 4.13 / 26.36  & 86.00 / 9.51 / 68.26 \\
DE (ensemble baseline) & 97.32 / 2.09 / 12.83 & 94.64 / 3.10 / 19.15  & \textbf{86.81} / 7.90 / 62.31 \\
\hline
BL + ERL & 97.28 / \textbf{1.35} / 12.09 & 94.97 / 3.21 / 17.31 & 85.77 / 6.75 / 58.02 \\
BL + ERL + MixUp & 97.28 / \underline{1.95} / 12.22 & 94.76 / \underline{2.12} / 16.31 & 86.07 / 5.13 / 56.32 \\ 

BL + ERL + MixUp + MCDrop & 97.32 / 2.76 / 12.13 & 94.66 / 2.15 / \underline{15.37} & 86.12 / 4.73 / 55.61 \\ 

BL + ERL + MixUp + MIMO & 97.36 / 2.04 / \underline{12.04} & \underline{95.01} / \underline{2.12} / 16.82 & 85.93 / \underline{4.69} / \underline{56.22} \\ 
BL + ERL + MixUp + DE & \textbf{97.44} / 2.78 / \textbf{11.45} & \textbf{95.31} / \textbf{1.56} / \textbf{14.24} & \underline{86.67} / \textbf{3.67} / \textbf{53.21}   \\ 
\hline

Train samples &  & \textbf{10 \%} & \\ \hline
RoBERTa (baseline) & 94.04 / 4.08 / 24.86 & 91.23 / 7.42 / 43.08 & 76.58 / 11.37 / 90.40 \\ 
DE (ensemble baseline) & \textbf{95.03} / 2.89 / \underline{19.02} & 91.44 / 4.88 / 29.51 & \underline{78.09} / 7.51 / \underline{78.96} \\
\hline
BL + ERL & 93.84 / 2.48 / 24.78 & 90.32 / 5.68 / 29.61 & 76.13 / 6.62 / 86.11 \\
BL + ERL + MixUp & 94.76 / \underline{2.23} / 22.02 & 90.89 / 4.52 / 26.59 & 76.25 / 6.39 / 84.20 \\ 

BL + ERL + MixUp + MCDrop & 94.68 / 2.41 / 21.92 & 90.93 / 4.26 / 26.16 & 76.16 / 4.69 / 82.54 \\ 

BL + ERL + MixUp + MIMO & 94.68 / \textbf{1.96} / 20.65 & \underline{91.75} / \underline{3.13} / \underline{23.96} & 76.89 / \underline{2.94} / 80.65 \\ 
BL + ERL + MixUp + DE & \underline{94.88} / 3.24 / \textbf{18.76} & \textbf{91.76 / 2.36 / 22.23} & \textbf{78.12} / \textbf{2.00} / \textbf{74.93} \\ 

\bottomrule 
\end{tabular}
\end{adjustbox}
\caption{ $\text{CALL}_{\text{MIMO}}$: BL+ERL+MixUp+MIMO. $\text{CALL}_{\text{DE}}$: BL+ERL+MixUp+DE.
The best and second best results are indicated in \textbf{bold} and \underline{underline}, respectively.
}
\label{tab:overall_tricks}
\end{table*}

\noindent \textbf{Diversity Analysis in Ensembles}.
Diversity of predictions in ensemble is one of the key factor of determining calibration performances \cite{MIMO}.
However, in the presence of overfitting, the diversity of predictions between ensemble members may decrease because the trained individual members would produce similar predictions that are overfitted to the same training data distribution \cite{NeuBoots}.

We hypothesize ensemble members of DE applied to PLMs may also suffer from overfitting.
Thus, we investigate whether the ensemble members are overfitted to NLL.
In Figure \ref{fig:de_nll}, DE trained with 10\% of the training data shows a different test NLL for each ensemble member, while DE trained with 100\% of the training data results in a closer NLL for the ensemble members as the training progresses.

According to our experimental result, members within the ensemble often fail to produce different predictions due to the overfitting, indicating that additional effective regularization schemes can be adopted to prevent overfitting when applying the ensemble to the PLM.
This finding also explains why ensemble techniques shows sub-par calibration performance compared to the regularization methods in the setting where full-data available.

\begin{figure}[th]
  \centering
    \begin{subfigure}[b]{0.23\textwidth}
         \centering
         \includegraphics[width=\textwidth, height=5.0cm]{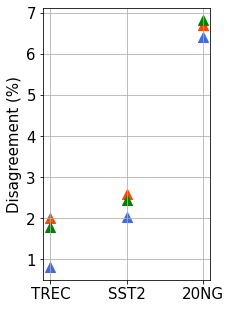}
         \subcaption{$D_{\text{train}}$ size: 100 \%}
     \end{subfigure}
    %  \hfill
     \begin{subfigure}[b]{0.23\textwidth}
         \centering
         \includegraphics[width=\textwidth, height=5.0cm]{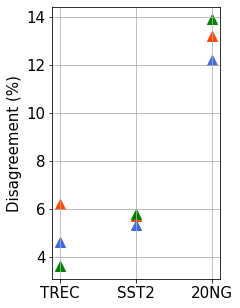}
         \subcaption{$D_{\text{train}}$ size: 10 \%}
     \end{subfigure}
  \caption{
  The diversity of predictions in ensemble with respect to the regularization methods.
  \textcolor{blue}{Blue}: DE; \textcolor{orange}{Orange}: DE+MixUp; \textcolor{darkgreen}{Green}: DE+BL+ERL. Results for MIMO and MCDrop are reported in Appendix \ref{sec:appendix_diversity}. A higher disagreement means that the models within the ensemble make different predictions.}
  \label{fig:DE_IMP}
\end{figure}

We investigate whether BL+ERL and MixUp methods can compensate for the aforementioned limitation of the ensemble method.
We measure disagreement score (see \citealp{MIMO}) to analyze the degree of diversity for predictions.
As shown in Figure \ref{fig:DE_IMP}, DE shows a high disagreement score in the low-resource regime.
When full-data are available, the disagreement score of DE is consistently the lowest for all datasets.
However, we observe that MixUp and BL+ERL significantly mitigate the reduction of predictive diversity for DE.

\section{Calibrated PLMs}\label{sec:call}

Through extensive analyses, we find that (1) MixUP that generate more diverse patterns helps improve the accuracy of BL+ERL, and (2) the reduced predictive diversity in the ensemble can be mitigated by BL+ERL and MixUp.

To this end, we report the calibration performance incrementally applying BL+ERL, MixUp, and ensemble techniques to the naive RoBERTa.
Specifically, we denote BL+ERL+MixUP+DE, and BL+ERL+MixUP+MIMO by $\text{CALL}_{\text{DE}}$, and $\text{CALL}_{\text{MIMO}}$, respectively.

In Table \ref{tab:overall_tricks}, overall, $\text{CALL}_{\text{DE}}$ achieves remarkable performance compared to DE on SST2 and 20NG datasets. 
$\text{CALL}_{\text{MIMO}}$ shows competitive performance with DE with respect to ECE and NLL.
This experiment shows that the calibration performance can be improved by the combinations using the ensemble, data augmentation, and confidence penalty losses in NLP tasks based on PLM, and each calibration method complements each other to further improve calibration performance without compromising accuracy.

\section{Conclusion}

In this work, we investigate the calibration effect of PLMs with various calibration methods applied.
As a result of a comprehensive analysis of how calibration methods work in PLMs, we find that (1) the confidence penalty losses have a trade-off between accuracy and calibration, and (2) ensemble techniques lose predictive diversity as training progresses, resulting in reduced calibration effectiveness.
To address these findings, we propose CALL, a combination of BL, ERL, MixUp, and ensemble learning. CALL reduces the risk of accuracy reduction through its data augmentation and ensemble techniques, and enhances the predictive diversity of ensemble methods by incorporating strong regularization and data augmentation. On multiple text classification datasets, CALL outperforms established baselines, making it a promising candidate as a strong baseline for calibration in text classification tasks.

\section*{Limitations}
Although the proposed framework achieves significantly improved calibration performance compared to the baselines, CALL still has room for performance improvement and may require more diverse approaches \cite{zadrozny2001obtaining,TEMPERATURE,FOCAL,SNGP}.
Another limitation is that we only address the ID calibration issue for PLMs.
Therefore, whether CALL could work well for out-of-distribution detection and generalization tasks is unclear.
We leave these questions for future research.

\section*{Ethics Statement}
The reliability of deep-learning models is crucial to the stable deployment of real-world NLP applications.
For example, the computer-aided resume recommendation system and neural conversational AI system should produce trustworthy predictions, because they are intimately related to the issue of trust in new technologies.
In this paper, through extensive empirical analysis, we address diverse calibration techniques and provide a detailed experimental guideline.
We hope our work will provide researchers with a new methodological perspective.

\section*{Acknowledgements}
This work was also supported by Research Fund (1.200086.01) of UNIST, Institute of Information \& communications Technology Planning \& Evaluation(IITP) grant funded by the Korea government(MSIT)(No. 2022-0-00612, Geometric and Physical Commonsense Reasoning based Behavior Intelligence for Embodied AI), and National Research Foundation of Korea(NRF) funded by the Korea government(MSIT)(2021R1C1C1009256).

\bibliography{anthology,custom}
\bibliographystyle{acl_natbib}

\clearpage

\appendix

\section{Computational Cost for Ensemble Methods}
\label{sec:computational_cost}

\begin{table}[th]
\begin{adjustbox}{width=7.5cm,center}
\begin{tabular}{c|ccc} \toprule
\begin{tabular}[c]{@{}c@{}}Latency $\downarrow$ (s) \\ (Train / Test)\end{tabular} & \textbf{TREC} & \textbf{SST2} & \textbf{20NG} \\ \hline
RoBERTa & 725.9 / 3.0 & 1031.9 / 8.2 & 1494.7 / 29.5 \\ \hline
MCDrop (M=2) & 725.9 / 5.8 & 1031.9 / 15.6 & 1494.7 / 58.8 \\
MIMO (M=2) & 840.7 / 3.5 & 1178.3 / 9.1 & 1720.0 / 34.0 \\
DE (M=2) & 1438.2 / 5.8 & 2060.7 / 15.6 & 3026.8 / 58.8 \\ \hline
$\text{CALL}_{\text{MIMO}}$ & 841.9 / 3.5 & 1180.2 / 9.1 &  1721.5 / 34.0 \\
$\text{CALL}_{\text{DE}}$ & 1440.3 / 5.8  & 2062.1 / 15.6 | & 3028.4 / 58.8 \\
\bottomrule 
\end{tabular}
\end{adjustbox}
\caption{Comparison of training/test time for ensemble approaches. We measure the computational time on an NVIDIA-V100 single GPU.}
\label{tab:computation}
\end{table}
Table \ref{tab:computation} includes computational costs for ensemble methods on a single GPU.
$\text{CALL}_{\text{DE}}$ (RoBERTa+BL+ERL+MixUP+DE) is almost the same as DE since only the
regularization term in the loss function and data augmentation process are added. Similarly, the
computation cost of $\text{CALL}_{\text{MIMO}}$ is almost the same as MIMO, and $\text{CALL}_{\text{MIMO}}$ achieves a significant speedup in training/test time compared to DE.

\section{Hyperparameter Setting}
\label{sec:hyperparameter_setting}
Selected hyperparameters are highlighted in bold.

\noindent \textbf{ERL}. Strength of the confidence penalty $\beta \in \{\textbf{0.001}, 0.005, 0.01, 0.1\}$. Empirically, PLMs trained with high beta (e.g., 0.1) showed sub-par classification accuracy. We
set the low beta as 0.001 for all experiments.

\noindent \textbf{LS}. $\epsilon$-smoothing parameter $\epsilon \in \{\textbf{0.01}, 0.05, 0.1\}$.

\noindent \textbf{EDA}. We follow the parameters recommended by the authors. Full-data setting: $\alpha=0.1$. Data scarcity setting: $\alpha=0.05$.
$\alpha$ is a parameter that indicates the percent of the words in a sentence that are changed.

\noindent \textbf{AEDA}. For each input sentence, $p=\{\textbf{5}, 10, 15\}$ percentage of the words are changed for low-resource regime, otherwise $p=\{5, \textbf{10}, 15\}$ words are changed.

\noindent \textbf{SR}. $p=\{\textbf{5}, 10, 15\}$ percentage of the words are changed for low-resource regime, otherwise $p=\{5, \textbf{10}, 15\}$ words are changed.

\noindent \textbf{MixUp}. $\alpha \in \{\textbf{0.1}, 0.5, 1.0\}$ (strength of interpolation).

\noindent \textbf{MCDrop}. $p \in \{0.01, 0.02, 0.03, 0.04, 0.05, 0.1\}$ is the Dropout rate. $M \in \{2, 3, 4, 5\}$. We choose the hyperparameters when the validation accuracy is best in each experiment.

\noindent \textbf{MIMO}. $M \in \{\textbf{2}, 3, 4, 5\}$. Validation accuracy tends to decrease when $M$ is increased. We choose input repetition parameter $p \in \{0.1, 0.2, 0.3, 0.4, 0.5\}$ when the validation accuracy is best in each experiment. Overall, $p=0.2$ is best.

\noindent \textbf{DE}. Full-data setting: $M \in \{\textbf{2}, 3, 4, 5\}$. 
Data scarcity setting: $M \in \{2, \textbf{3}, 4, 5\}$.

\section{Empirical Result for BERT}
\label{sec:bert_result}
We report empirical results for BERT in Table \ref{tab:overall_bert} and Table \ref{tab:overall_bert_scarcity}.

\begin{figure}[th]
  \centering
    \begin{subfigure}[b]{0.23\textwidth}
         \centering
         \includegraphics[width=\textwidth, height=5.0cm]{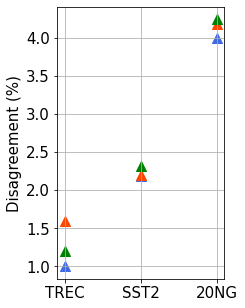}
         \subcaption{$D_{\text{train}}$ size: 100 \%}
     \end{subfigure}
    %  \hfill
     \begin{subfigure}[b]{0.23\textwidth}
         \centering
         \includegraphics[width=\textwidth, height=5.0cm]{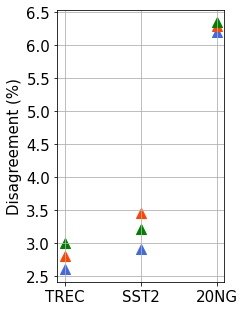}
         \subcaption{$D_{\text{train}}$ size: 10 \%}
     \end{subfigure}
  \caption{Effect of regularization with respect to diversity of predictions in ensemble. \textcolor{blue}{Blue}: MCDrop; \textcolor{orange}{Orange}: MCDrop+MixUp; \textcolor{darkgreen}{Green}: MCDrop+BL+ERL.}
  \label{fig:MCDROP_IMP}
\end{figure}

\begin{figure}[th]
  \centering
    \begin{subfigure}[b]{0.23\textwidth}
         \centering
         \includegraphics[width=\textwidth, height=5.0cm]{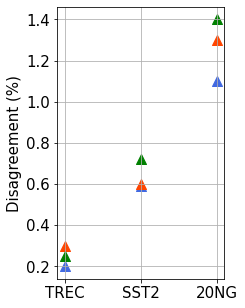}
         \subcaption{$D_{\text{train}}$ size: 100 \%}
     \end{subfigure}
    %  \hfill
     \begin{subfigure}[b]{0.23\textwidth}
         \centering
         \includegraphics[width=\textwidth, height=5.0cm]{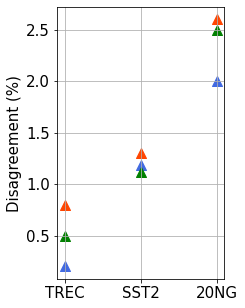}
         \subcaption{$D_{\text{train}}$ size: 10 \%}
     \end{subfigure}
  \caption{Effect of regularization with respect to diversity of predictions in ensemble. \textcolor{blue}{Blue}: MIMO; \textcolor{orange}{Orange}: MIMO+MixUp; \textcolor{darkgreen}{Green}: MIMO+BL+ERL.}
  \label{fig:MIMO_IMP}
\end{figure}

\section{Analysis Diversity}
\label{sec:appendix_diversity}
We report diversity measure for MCDrop and MIMO in Figure \ref{fig:MCDROP_IMP} and Figure \ref{fig:MIMO_IMP}, respectively.

\begin{table*}[h]
\begin{adjustbox}{width=15cm,center}
\begin{tabular}{l|ccc} \toprule
Acc$\uparrow$ / ECE$\downarrow$ / NLL$\downarrow$ & \textbf{TREC} & \textbf{SST2} & \textbf{20NG} \\ \hline
BERT (baseline) & 97.24 / 2.44 / 13.20 & 91.26 / 5.19 / 33.77  & 85.45 / 9.98 / 70.33 \\
CE+ERL & 97.24 / 2.43 / 13.18 & 91.23 / 5.15 / 33.66  & 85.45 / 10.26 / 71.58 \\
CE+LS & 97.11 / 2.08 / 12.22 & 91.50 / 5.09 / 26.80  & 85.39 / 6.42 / 60.39 \\
BL & 97.64 / \textbf{1.29} / 10.38 & 91.33 / 5.18 / 28.01  & 85.28 / 7.25 / 60.46 \\
BL+ERL & 96.76 / 1.42 / 12.17 & 91.29 / 4.99 / 26.66  & 85.36 / 6.58 / 59.25 \\
BL+LS & 97.13 / 1.48 / 12.09 & 91.07 / 4.85 / 27.02  & 85.20 / 6.99 / 60.14 \\
\hline
SR & 97.48 / 1.96 / 10.37 & 91.83 / 5.11 / 29.53  & 85.50 / 9.60 / 68.14 \\
AEDA & 97.60 / 1.60 / 10.57 &  91.54 / 7.23 / 43.63 & 85.49 / 9.76 / 68.87 \\
EDA & 97.56 / 1.59 / 10.58 & 91.63 / 3.44 / 23.63  & 85.47 / 9.23 / 65.66 \\
MixUp & 97.40 / 1.30 / 11.12 & 91.66 / 5.89 / 28.78  & 85.63 / 8.92 / 66.20 \\
\hline
MCDrop & 97.32 / 2.08 / 12.97 & 91.52 / 5.89 / 31.28 & 85.35 / 9.90 / 68.56 \\
MIMO & 97.44 / 1.63 / 10.68 &  91.40 / 6.25 / 32.14 & 85.37 / 8.68 / 62.82 \\
DE & 97.32 / 1.98 / 11.26 & \textbf{91.92} / 4.14 / 27.27  & 85.86 / 7.99 / 62.81 \\
\hline
BL+ERL+MixUp+MCDrop & 97.34 / 2.01 / 12.37 & 91.59 / 3.61 / 28.54  & 85.37 / 5.62 / 60.18 \\
BL+ERL+MixUp+MIMO ($\text{CALL}_{\text{MIMO}}$) & 97.56 / 1.52 / 10.40 & 91.37 / 5.03 / 25.96  & 85.33 / 4.87 / 58.06 \\
BL+ERL+MixUp+DE ($\text{CALL}_{\text{DE}}$) & \textbf{97.79} / 2.82 / \textbf{10.18} & 91.82 / \textbf{2.58} / \textbf{22.19}  & \textbf{86.05} / \textbf{3.62} / \textbf{54.03} \\
\bottomrule 
\end{tabular}
\end{adjustbox}
\caption{Result for BERT with diverse calibration techniques. The best results are indicated in \textbf{bold}.}
\label{tab:overall_bert}
\end{table*}

\begin{table*}[h]
\begin{adjustbox}{width=15cm,center}
\begin{tabular}{l|ccc} \toprule
Acc$\uparrow$ / ECE$\downarrow$ / NLL$\downarrow$ & \textbf{TREC} & \textbf{SST2} & \textbf{20NG} \\ \hline
BERT (baseline) & 93.40 / 4.43 / 25.16 &  87.47 / 9.49 / 52.36 & 73.79 / 10.90 / 96.02 \\
CE+ERL & 93.40 / 4.40 / 25.13 & 87.48 / 9.50 / 51.66  & 73.77 / 10.84 / 95.96 \\
CE+LS & 93.28 / 3.87 / 24.13 &  87.44 / 7.99 / 37.05 & 73.57 / 8.07 / 94.78 \\
BL & 93.60 / \textbf{2.33} / 21.54 & 87.26 / 7.25 / 38.74  & 73.96 / 6.63 / 91.02 \\
BL+ERL & 93.25 / 2.38 / 21.95 & 87.56 / 6.83 / 36.96  & 74.21 / 5.63 / 90.94 \\
BL+LS & 93.14 / 2.41 / 22.03 & 87.78 / 6.01 / 36.76 & 73.91 / 5.89 / 92.37 \\
\hline
SR & 92.52 / 4.67 / 28.37 & 87.74 / 8.62 / 46.59  & 74.00 / 10.93 / 95.34 \\
AEDA & 93.44 / 4.36 / 24.48 & 87.71 / 9.03 / 48.55  & 73.65 / 11.52 / 97.43 \\
EDA & 91.88 / 4.30 / 28.30 & 87.44 / 8.93 / 44.94  & 74.04 / 10.33 / 94.26 \\
MixUp & 93.88 / 2.76 / 20.47 & 87.65 / 7.20 / 37.47  & 74.01 / 9.04 / 95.31 \\
\hline
MCDrop & 93.56 / 3.53 / 24.89 & 87.43 / 8.87 / 50.13  & 73.81 / 10.24 / 94.77 \\
MIMO & 93.88 / 2.62 / 21.53 &  87.55 / 6.09 / 34.82 & 73.80 / 7.25 / 88.65 \\
DE  & 93.68 / 2.91 / 21.13 & 87.92 / 6.76 / 38.44  & 75.19 / 7.52 / 85.81 \\
\hline
BL+ERL+MixUp+MCDrop & 93.45 / 3.51 / 23.77 & 87.58 / 5.42 / 34.31  & 73.80 / 7.35 / 90.69 \\
BL+ERL+MixUp+MIMO ($\text{CALL}_{\text{MIMO}}$) & 93.56 / 2.91 / 21.20 &  87.70 / 5.85 / 34.35 & 74.11 / 5.21 / 89.93 \\
BL+ERL+MixUp+DE ($\text{CALL}_{\text{DE}}$) & \textbf{94.24} / 3.41 / \textbf{19.79} &  \textbf{88.25} / \textbf{2.48} / \textbf{28.65} & \textbf{75.68} / \textbf{2.20} / \textbf{82.90} \\
\bottomrule 
\end{tabular}
\end{adjustbox}
\caption{Result for BERT with diverse calibration techniques on the low-resource regime. The best results are indicated in \textbf{bold}.}
\label{tab:overall_bert_scarcity}
\end{table*}

\end{document}